\documentclass[conference]{IEEEtran}
\IEEEoverridecommandlockouts
\usepackage{cite}
\usepackage{amsmath,amssymb,amsfonts}
\usepackage{algorithmic}
\usepackage{graphicx}
\usepackage{textcomp}
\usepackage{xcolor}
\def\BibTeX{{\rm B\kern-.05em{\sc i\kern-.025em b}\kern-.08em
    T\kern-.1667em\lower.7ex\hbox{E}\kern-.125emX}}
    
\usepackage{multirow}
\usepackage{booktabs}
\usepackage{subcaption}
\usepackage{caption}
\usepackage{threeparttable}

\usepackage{colortbl, soul}

\usepackage{color}

\begin{document}

\title{Learning Proxemic Behavior Using Reinforcement Learning with Cognitive Agents}

\author{\IEEEauthorblockN{Cristian Millán-Arias$^1$, Bruno Fernandes$^1$, Francisco Cruz$^{2,3}$}
\IEEEauthorblockA{$^1$Escola Politécnica de Pernambuco, 
Universidade de Pernambuco, Recife-PE, Brazil\\
\IEEEauthorblockA{$^2$School of Information Technology, Deakin University, Geelong, Australia\\
\IEEEauthorblockA{$^3$Escuela de Ingeniería, Universidad Central de Chile, Santiago, Chile\\
Emails: \{ccma, bjtf\}@ecomp.poli.br, francisco.cruz@deakin.edu.au} 
}
}
}

\iffalse
\author{\IEEEauthorblockN{Cristian Millán-Arias}
\IEEEauthorblockA{\textit{Escola Politécnica de Pernambuco,} \\
\textit{Universidade de Pernambuco}\\
Recife-PE, Brazil \\
ccma@ecomp.poli.br}
\and
\IEEEauthorblockN{Bruno Fernandes}
\IEEEauthorblockA{\textit{Escola Politécnica de Pernambuco,} \\
\textit{Universidade de Pernambuco}\\
Recife-PE, Brazil \\
bjtf@ecomp.poli.br}
\and
\IEEEauthorblockN{Francisco Cruz}
\IEEEauthorblockA{\textit{School of Information Technology} \\
\textit{Deakin University}\\
Geelong, Australia}
\IEEEauthorblockA{\textit{Escuela de Ingeniería, Universidad} \\
\textit{Central de Chile}\\
Santiago, Chile \\
francisco.cruz@deakin.edu.au}
}
\fi

\maketitle

\begin{abstract}
Proxemics is a branch of non-verbal communication concerned with studying the spatial behavior of people and animals.
This behavior is an essential part of the communication process due to delimit the acceptable distance to interact with another being.
With increasing research on human-agent interaction,
new alternatives are needed that allow optimal communication, avoiding agents feeling uncomfortable.
Several works consider proxemic behavior with cognitive agents, where human-robot interaction techniques and machine learning are implemented. 
However, environments consider fixed personal space and that the agent previously knows it.
In this work, we aim to study how agents behave in environments based on proxemic behavior, and propose a modified gridworld to that aim. 
This environment considers an issuer with proxemic behavior that provides a disagreement signal to the agent. 
Our results show that the learning agent can identify the proxemic space when the issuer gives feedback about agent performance.
 
\end{abstract}


\section{Introduction}
Proxemics is the study of spatial behavior, concerned with territoriality, interpersonal distance, spatial arrangements, crowding, and others aspects of the physical environment that affect behavior. 
The term was coined by Hall et al.~\cite{hall1968proxemics}, when he proposes a fixed measure of personal space, a set of regions around a person to delimit the acceptable distance to interact with other people.
In recent years, human-agent interaction has taken hold in the scientific community. 
Furthermore, the humanization of agents is an expected event, given technological advances and human nature.
Thus, an optimal interaction is necessary on both the agent and the person \cite{churamani2020icub}.

Reinforcement Learning (RL) is a learning paradigm that tries to solve the problem of an agent interacting with the environment to learn the desired task autonomously \cite{Sutton2018}. 
The agent must sense a state from the environment and take actions that affect it to reach a new state. 
The agent receives a reward signal from the environment that it tries to maximize throughout the learning for each action taken.
The agent takes actions from its own experience, or can be guided by an external trainer that provides feedback \cite{Millan2019, millan2021robust}.
Proxemic behavior has been used in different areas with cognitive agents. 
For example, in human-robot interaction, it has been studied how people behave in the presence of an artificial agent or robot and how their perception of the personal space of the agent has been \cite{mumm2011human, eresha2013investigating}.
Moreover, in machine learning, it has been studied how artificial agents sense the personal space of other cognitive agents, and it has been to identify and learn personal space \cite{patompak2020learning, mitsunaga2006robot}.

\section{Proxemic Behavior in Cognitive Agents}

\iffalse
\subsection{Q-learning Reinforcement Learning}

The off-policy Q-learning \cite{watkins1992q} is a model-free algorithm to learn the value of an action in a particular state. 
Moreover, it is one of the most important breakthroughs in RL. 
The core of the algorithm is a Bellman equation, where the value $\max_{a} Q(s_{t+1}, a)$ is used to update of the value $Q(s_t,a)$.  
Thus, the state-action value function for discounted returns are update as follows:

\begin{equation*}
    Q(s_t,a) \longleftarrow Q(s_t,a) + \alpha \left( \rho_t + \gamma \max_{a} Q(s_{t+1}, a) - Q(s_t,a) \right),
\end{equation*}

\noindent where $(s_t, a)$ is the state-action pair, $\alpha$ the learning rate, and $\gamma$ the discounted factor.
\fi

% \subsection{Experimental Environment}

We propose a modified version of the gridworld problem. 
In this environment, an issuer is placed on one fixed state and is responsible for giving a signal of disagreement when the learning agent is too close.
Two regions are defined around the issuer, the uncomfortable region and the target region.
The uncomfortable region is those states that make up a square around the issuer. 
In this work, the uncomfortable region is only an area of negative reward. 
Similarly, the target region is those states that make up a square around the uncomfortable region.
A new action, PING, is added to the four traditional ones (UP, DOWN, LEFT, RIGHT). 
This action represents a communication signal with the issuer, i.e., the agent send a ping to ask the issuer if it is in the target region.
The task finishes in three conditions:
\begin{itemize}
    \item \textit{Condition C1}: When the agent reaches the issuer.
    \item \textit{Condition C2}: When the agent sends ping five times out of the target region.
    \item \textit{Condition C3}: When the agent sends ping in the target region.
\end{itemize}

The reward function is defined as:
\begin{equation}
    \rho = \rho_{issuer} + \rho_{grid},
\end{equation}

\noindent where $\rho_{issuer}$ is a numerical reward given by the issuer when the agent performs the PING action, and $\rho_{grid}$ is the environment reward defined as:

\begin{equation}
\rho_{grid} =\left\{  
\begin{array}{rcl}
	-1.0 &\mbox{if} & \mbox{conditions C1 or C2} \\
	-0.8 & \mbox{if}& \mbox{reach to uncomfortable region} \\
	-0.4 & \mbox{if} & \mbox{give incorrectly ping} \\
	-0.1 & \mbox{if} & \mbox{another state} \\
	+1.0 & \mbox{if}& \mbox{condition C3} \\
\end{array} \right. .
\end{equation}

In our experiments, we use a $10\times 12$ gridworld, the issuer is placed in $(6, 8)$ and maintains fixed during the training. 
Each agent starts the training in the superior left corner of the grid $(0, 0)$.

\section{experimental results}

To explore the behavior of agents in a proxemic environment, we apply the Q-Learning algorithm \cite{watkins1992q} in the modified gridworld problem. 
We use $\epsilon$-greedy to select random actions. 
We set the values of $\epsilon$ and learning rate $\alpha$ at 0.6 and the discount factor $\gamma$ at 0.9.
In our experiments, $100$ agents are trained with $10000$ time-steps.

In our experiments, we consider three scenarios:
\begin{itemize}
    \item Scenario S1: When $\rho_{issuer}=0$ in each PING action selection.
    \item Scenario S2: When $\rho_{issuer}$ is a random value in $(-1,1)$.
    \item Scenario S3: When $\rho_{issuer}=-\dfrac{d_1(P_{agent}, P_{issuer})}{d_1(P_{start}, P_{issuer})}$, where $d_1(\cdot, \cdot)$ is the $L_1$ distance, $P_{agent}$, and $P_{issuer}$, and $P_{start}$ are the position on the grid from the agent, the issuer and the start point.
    This value indicates that states farthest from the issuer have a low reward, and the denominator standardizes it on (-1, 0).
\end{itemize}

\begin{figure}[!t]
    \centering
    \subcaptionbox{Scenario S1\label{results:a}}{\includegraphics[width=2.8cm]{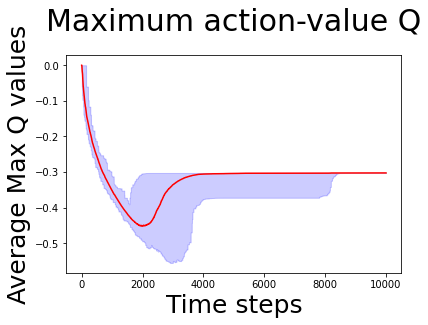}}\hspace{0.3em}
    \subcaptionbox{Scenario S2\label{results:b}}{\includegraphics[width=2.8cm]{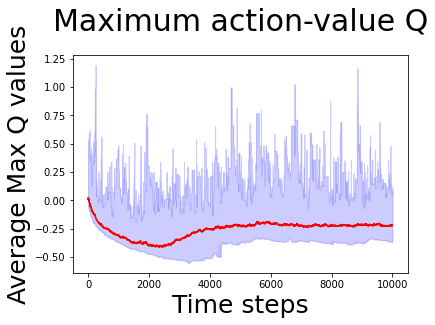}}\hspace{0.3em}
    \subcaptionbox{Scenario S3\label{results:c}}{\includegraphics[width=2.8cm]{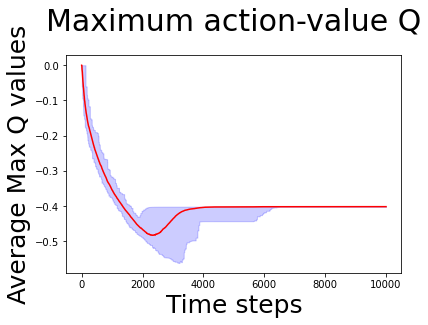}}%
    
\bigskip

    \subcaptionbox{Scenario S1\label{results:d}}{\includegraphics[width=2.8cm]{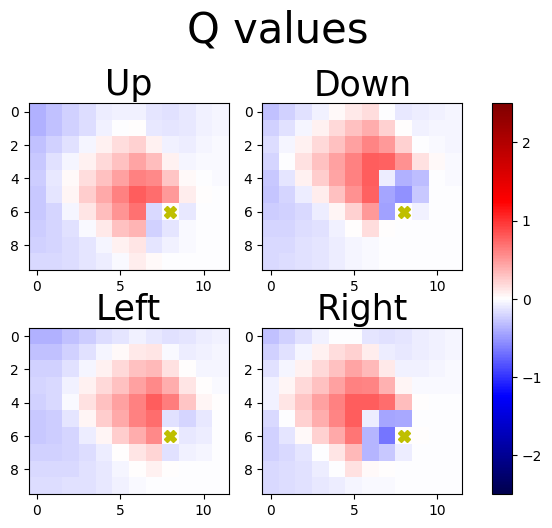}}\hspace{0.3em}
    \subcaptionbox{Scenario S2\label{results:e}}{\includegraphics[width=2.8cm]{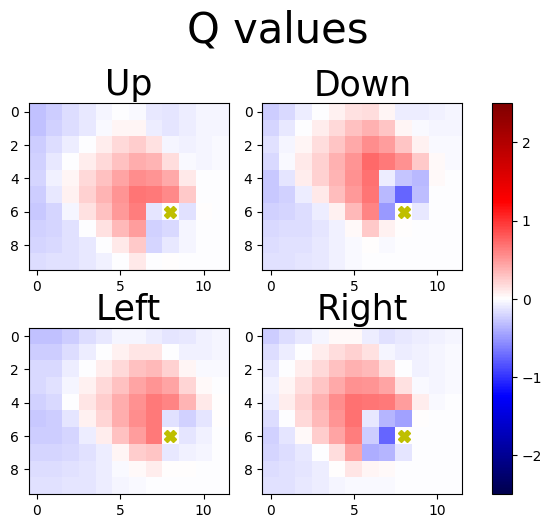}}\hspace{0.3em}
    \subcaptionbox{Scenario S3\label{results:f}}{\includegraphics[width=2.8cm]{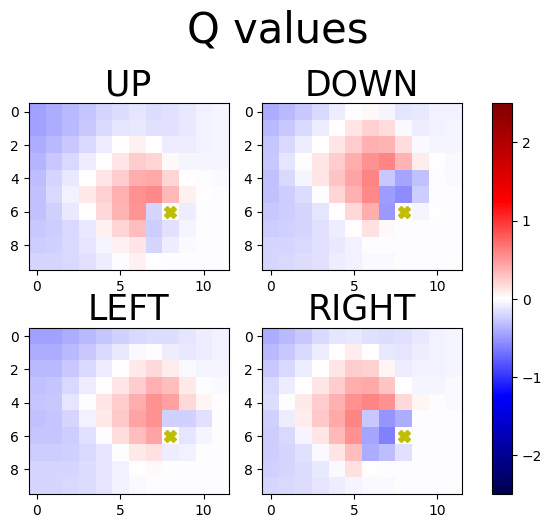}}%
    
\bigskip

    \subcaptionbox{Scenario S1\label{results:g}}{\includegraphics[width=2.8cm]{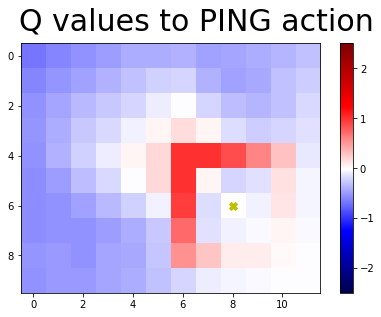}}\hspace{0.3em}
    \subcaptionbox{Scenario S2\label{results:h}}{\includegraphics[width=2.8cm]{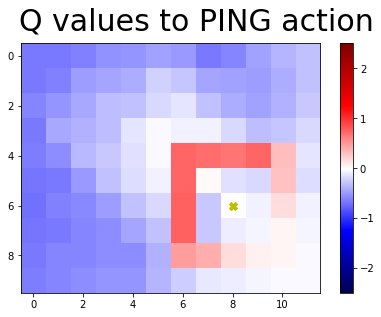}}\hspace{0.3em}
    \subcaptionbox{Scenario S3\label{results:i}}{\includegraphics[width=2.8cm]{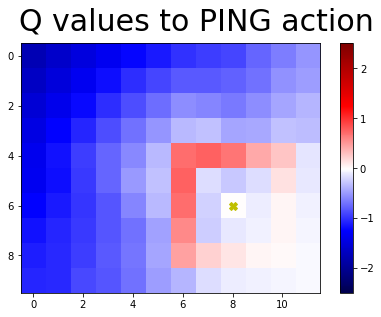}}%
    \caption{Results of our experiments in three scenarios. The first row (\ref{results:a}-\ref{results:c}) shows the average of maximum Q values per time step, the shaded area is between the maximum and minimum Q values from 100 agents. The second row (\ref{results:d})-\ref{results:f}) shows the average of final Q value for the actions UP, DOWN, LEFT, and RIGHT. The last row (\ref{results:g}-\ref{results:i}) show the average final Q values for the PING action.}
    \label{results}
\end{figure}

Fig. \ref{results} shows the results of our experiments in the three scenarios. 
In terms of maximum Q-value per time step (Fig. \ref{results:a}-\ref{results:c}), there exists more variability when the issuer gives random rewards. 
However, according to the other scenarios, the maximum Q-value reaches values around $1$ in several time steps. 
On the other hand, when the issuer gives rewards based on distance, the variability of Q-values is minor.

The Q values for each action are more focused on the target region than in the other scenarios when the issuer gives rewards based on distance (Fig. \ref{results:d}-\ref{results:f}). 
This is expected due to the agent receives additional information on how to move through the grid.

Concerning the PING, the issuer reward highlights the importance of giving ping in the target region, as shown in Fig. \ref{results:g}-\ref{results:i}.
The graph shows how the states of the target region have only Q-values greater than zero. 
While with the lack of information (Scenario S1), states outside the target region have Q-values greater than zero. 

\iffalse
The first row (Fig. \ref{results:a}-\ref{results:c}) shows the average maximum Q-value per time step. 
It is clear that exists more variability when the issuer gives random rewards. 
However, according to the other scenarios, the maximum Q-value reaches values around $1$ in several time steps. 
On the other hand, when the issuer gives rewards based on distance, the variability of Q-values is minor.

The second row (Fig. \ref{results:d}-\ref{results:f}) shows the average final Q-value for each action.
The graph shows that the upper left corner of the external region has Q-values greater than zero. 
This behavior occurs because the agent always starts in the same position.
Regarding the scenario, the Q-values when the issuer gives rewards based on distance are more focused on the region than in the other scenarios. 
This is expected due to the agent receives additional information on how to move through the grid. 
It also explains the low variability with the Q-values. 
In contrast, when the issuer gives random rewards, the Q-values are more dispersed due to the lack of helpful information.

Concerning the PING, the issuer reward highlights the importance of giving ping in the correct region, as shown in the last row of Fig. \ref{results:g}-\ref{results:i}.
The graph shows how the states of the external region have only Q-values greater than zero. 
While with the lack of information (scenario S1), states outside the region have Q-values greater than zero. 
\fi

\section{Conclusions and future alignments}

In this paper, we study the agent performance in an environment based on proxemic behavior. 
We implement a modified gridworld problem, where an issuer agent gives a signal of disagreement when the RL agent performs a ping action. 
Moreover, the target is a proxemic region around the issuer instead of the traditional goal state. 
In our experiments, we consider three scenarios based on the reward type that the issuer gives. 
Our results show that the agent can reach the target region, even when the issuer does not give information. 
On the other hand, the issuer reward gives more information about performing the ping action, even when random information is given. 
Thus, the agent can identify the correct region to give a ping signal.

In our implementation, we consider that the proxemic region is symmetric around the user. 
Thus, in our future alignments, we intend to implement asymmetric proxemic regions to mimic the human proxemic behavior. 
Also, we contemplate implementing no fixed proxemic region, considering that the proxemic space changes by external factors.
Finally, we intend to study the agent performance in more complex environments, involving other algorithms and techniques of reinforcement learning and deep learning.

\section*{Acknowledgment}
We would like to gratefully acknowledge financing in part by the Coordenac\~ao de Aperfei\c coamento de Pessoal de N\'ivel Superior-Brasil (CAPES) - Finance Code 001, and the Brazilian agencies FACEPE and CNPq - Code 432818$/$2018-9.

\bibliographystyle{IEEEtran}
\bibliography{bibliography}

\begin{thebibliography}{10}
\providecommand{\url}[1]{#1}
\csname url@samestyle\endcsname
\providecommand{\newblock}{\relax}
\providecommand{\bibinfo}[2]{#2}
\providecommand{\BIBentrySTDinterwordspacing}{\spaceskip=0pt\relax}
\providecommand{\BIBentryALTinterwordstretchfactor}{4}
\providecommand{\BIBentryALTinterwordspacing}{\spaceskip=\fontdimen2\font plus
\BIBentryALTinterwordstretchfactor\fontdimen3\font minus
  \fontdimen4\font\relax}
\providecommand{\BIBforeignlanguage}[2]{{%
\expandafter\ifx\csname l@#1\endcsname\relax
\typeout{** WARNING: IEEEtran.bst: No hyphenation pattern has been}%
\typeout{** loaded for the language `#1'. Using the pattern for}%
\typeout{** the default language instead.}%
\else
\language=\csname l@#1\endcsname
\fi
#2}}
\providecommand{\BIBdecl}{\relax}
\BIBdecl

\bibitem{hall1968proxemics}
E.~T. Hall, R.~L. Birdwhistell, B.~Bock, P.~Bohannan, A.~R. Diebold~Jr,
  M.~Durbin, M.~S. Edmonson, J.~Fischer, D.~Hymes, S.~T. Kimball \emph{et~al.},
  ``Proxemics [and comments and replies],'' \emph{Current anthropology},
  vol.~9, no. 2/3, pp. 83--108, 1968.

\bibitem{churamani2020icub}
N.~Churamani, F.~Cruz, S.~Griffiths, and P.~Barros, ``icub: learning emotion
  expressions using human reward,'' \emph{arXiv preprint arXiv:2003.13483},
  2020.

\bibitem{Sutton2018}
R.~S. Sutton and A.~G. Barto, \emph{{Reinforcement Learning: An
  Introduction}}.\hskip 1em plus 0.5em minus 0.4em\relax Cambridge,
  Massachusetts: MIT press, 2018.

\bibitem{Millan2019}
C.~Mill{\'{a}}n, B.~Fernandes, and F.~Cruz, ``{Human feedback in continuous
  actor-critic reinforcement learning},'' in \emph{Proceedings European
  Symposium on Artificial Neural Networks, Computational Intelligence and
  Machine Learning}, no. April, Bruges (Belgium), 2019, pp. 661--666.

\bibitem{millan2021robust}
C.~Millan-Arias, B.~Fernandes, F.~Cruz, R.~Dazeley, and S.~Fernandes, ``A
  robust approach for continuous interactive actor-critic algorithms,''
  \emph{{IEEE} Access}, vol.~9, pp. 104\,242--104\,260, 2021.

\bibitem{mumm2011human}
J.~Mumm and B.~Mutlu, ``Human-robot proxemics: physical and psychological
  distancing in human-robot interaction,'' in \emph{Proceedings of the 6th
  international conference on Human-robot interaction}, 2011, pp. 331--338.

\bibitem{eresha2013investigating}
G.~Eresha, M.~H{\"a}ring, B.~Endrass, E.~Andr{\'e}, and M.~Obaid,
  ``Investigating the influence of culture on proxemic behaviors for humanoid
  robots,'' in \emph{2013 IEEE Ro-Man}.\hskip 1em plus 0.5em minus 0.4em\relax
  IEEE, 2013, pp. 430--435.

\bibitem{patompak2020learning}
P.~Patompak, S.~Jeong, I.~Nilkhamhang, and N.~Y. Chong, ``Learning proxemics
  for personalized human--robot social interaction,'' \emph{International
  Journal of Social Robotics}, vol.~12, no.~1, pp. 267--280, 2020.

\bibitem{mitsunaga2006robot}
N.~Mitsunaga, C.~Smith, T.~Kanda, H.~Ishiguro, and N.~Hagita, ``Robot behavior
  adaptation for human-robot interaction based on policy gradient reinforcement
  learning,'' \emph{Journal of the Robotics Society of Japan}, vol.~24, no.~7,
  pp. 820--829, 2006.

\bibitem{watkins1992q}
C.~J. Watkins and P.~Dayan, ``Q-learning,'' \emph{Machine learning}, vol.~8,
  no. 3-4, pp. 279--292, 1992.

\end{thebibliography}

\end{document}